\documentclass[conference]{IEEEtran}
\usepackage{cite}
\usepackage{amsmath,amssymb,amsfonts}
\usepackage{graphicx}
\usepackage{booktabs}
\usepackage{array}
\usepackage{url}
\usepackage[table]{xcolor}
\usepackage{enumitem}
\usepackage{hyperref}

\def\BibTeX{{\rm B\kern-.05em{\sc i\kern-.025em b}\kern-.08em T\kern-.1667em\lower.7ex\hbox{E}\kern-.125emX}}

\definecolor{catCorrectness}{RGB}{232,240,248}   
\definecolor{catPerformance}{RGB}{255,245,230}   
\definecolor{catCode}{RGB}{236,247,233}          
\definecolor{catAlgorithm}{RGB}{242,236,248}     
\definecolor{catPETSc}{RGB}{248,232,235}         
\definecolor{compositeTint}{RGB}{240,240,240}    

\newcommand{\petscbench}{\textsc{petscagent-bench}}

\newcommand{\modellogo}[1]{%
  \raisebox{-0.2ex}{\includegraphics[height=1.6ex]{#1}}}

\usepackage{color}
\definecolor{linkblue}{RGB}{0,0,140}
\hypersetup{
     colorlinks=true,
     citecolor=linkblue,
     filecolor=black,
     linkcolor=linkblue,
     urlcolor=linkblue
}
\usepackage{doi}

\begin{document}

\title{An Agentic Evaluation Framework for AI-Generated Scientific Code in PETSc}

\author{%
\IEEEauthorblockN{Hong Zhang\IEEEauthorrefmark{1},
                  Barry Smith\IEEEauthorrefmark{2},
                  Satish Balay\IEEEauthorrefmark{1},
                  Le Chen\IEEEauthorrefmark{1},
                  Murat Keceli\IEEEauthorrefmark{1},
                  Lois Curfman McInnes\IEEEauthorrefmark{1},
                  Junchao Zhang\IEEEauthorrefmark{1}}
\IEEEauthorblockA{\IEEEauthorrefmark{1}Argonne National Laboratory, Lemont, IL, USA\\
\{hongzhang, balay, lechen, keceli, mcinnes, jczhang\}@anl.gov}
\IEEEauthorblockA{\IEEEauthorrefmark{2}bsmith@petsc.dev}
}

\maketitle

\begin{abstract}
While LLMs have accelerated scientific code
generation, comprehensively evaluating generated code remains  challenging.
Many benchmarks emphasize functional correctness or task completion, which is
insufficient for code built on production HPC libraries, where solver selection, API conventions,
memory management, parallel awareness, and performance also matter.
We introduce \petscbench{}, a multidimensional benchmark and agent-based framework for assessing whether AI-generated scientific code uses a production HPC library as an expert would.
A tool-augmented
evaluator compiles, executes, and measures code and combines deterministic checks with LLM-based assessments in a 14-evaluator pipeline spanning
five categories: correctness, performance, code quality, algorithmic
appropriateness, and library-specific conventions.
A2A and MCP 
enable black-box evaluation of compatible coding agents. 
Across realistic PETSc problems, frontier models 
generate readable, well-structured code but struggle with correctness on challenging problems and with
library-specific conventions even when code compiles and runs---limitations that conventional pass/fail evaluation does not capture.

\end{abstract}

\begin{IEEEkeywords}
Large language models, code generation, scientific computing, HPC, PETSc, agent-based evaluation, benchmarking
\end{IEEEkeywords}

\section{Introduction}
\label{sec:intro}

Production scientific software is built on numerical libraries.
Packages such as PETSc~\cite{petsc-user-ref} and OpenFOAM~\cite{greenshields2025} 
provide solvers, data structures, and interfaces developed over decades of research in numerical methods and parallel computing.
Using them well is itself a demanding skill.
A practitioner must know which components suit a given problem and how to compose them according to each library's conventions.

Large language models (LLMs) are increasingly used to write such code, with demonstrated results in code generation~\cite{chen2021evaluating,nichols2024pareval}, code translation~\cite{chen2025beyond}, and simulation workflow automation~\cite{zhang2025mooseagent}; they are also now deployed as autonomous coding agents.
The capability that matters for an LLM in this setting is using a library correctly. A capable model selects appropriate solvers and data structures for the problem, composes API calls in the prescribed patterns, and follows the library's conventions for error handling and resource management.

Many code-generation benchmarks emphasize functional correctness or task completion (Section~\ref{sec:related}).
Many prompt an LLM to produce code and check it against test cases, often reducing the result to a pass/fail or pass@k score.
HPC-oriented benchmarks may additionally incorporate compilation, parallel execution, performance, or numerical accuracy, but generally do not assess whether a production numerical library is used as an expert would.
For library-based code, \emph{passing tests is necessary but not sufficient}.
A representative case from our experiments makes this concrete.
In our experiments, frontier models often generate PETSc code that compiles, runs, and produces the correct answer, yet use the legacy \texttt{CHKERRQ} macro rather than \texttt{PetscCall} for error handling, even though PETSc deprecated \texttt{CHKERRQ} in version~3.17 (2022).
A pass/fail benchmark records such a submission as a complete success, while a PETSc developer would see at once that it does not follow current library conventions.
What separates production-quality library code from merely working code is a set of properties these benchmarks do not measure.
They include solver appropriateness, idiomatic API use, proper error handling, and efficient use of the library's abstractions.

Measuring these properties calls for a different kind of evaluator.
Two common evaluation approaches illustrate the complementary capabilities required: deterministic testing and LLM-based assessment. Neither is adequate on its own for the setting considered here.
Static test scripts are reproducible and objective, but conventional correctness tests primarily establish whether code compiles, executes, and matches expected output.
Such tests do not by themselves determine whether a call sequence is idiomatic or whether the chosen solver suits the problem.
LLM-as-a-judge methods~\cite{zheng2024judging} assess style and semantics, but a text-only judge does not directly 
compile code against a library, run a binary in parallel, or measure execution time.
Library-based scientific code requires both capabilities at once: execution and expert judgment.

\begin{figure*}[!tb]
  \centering
  \includegraphics[width=0.88\textwidth]{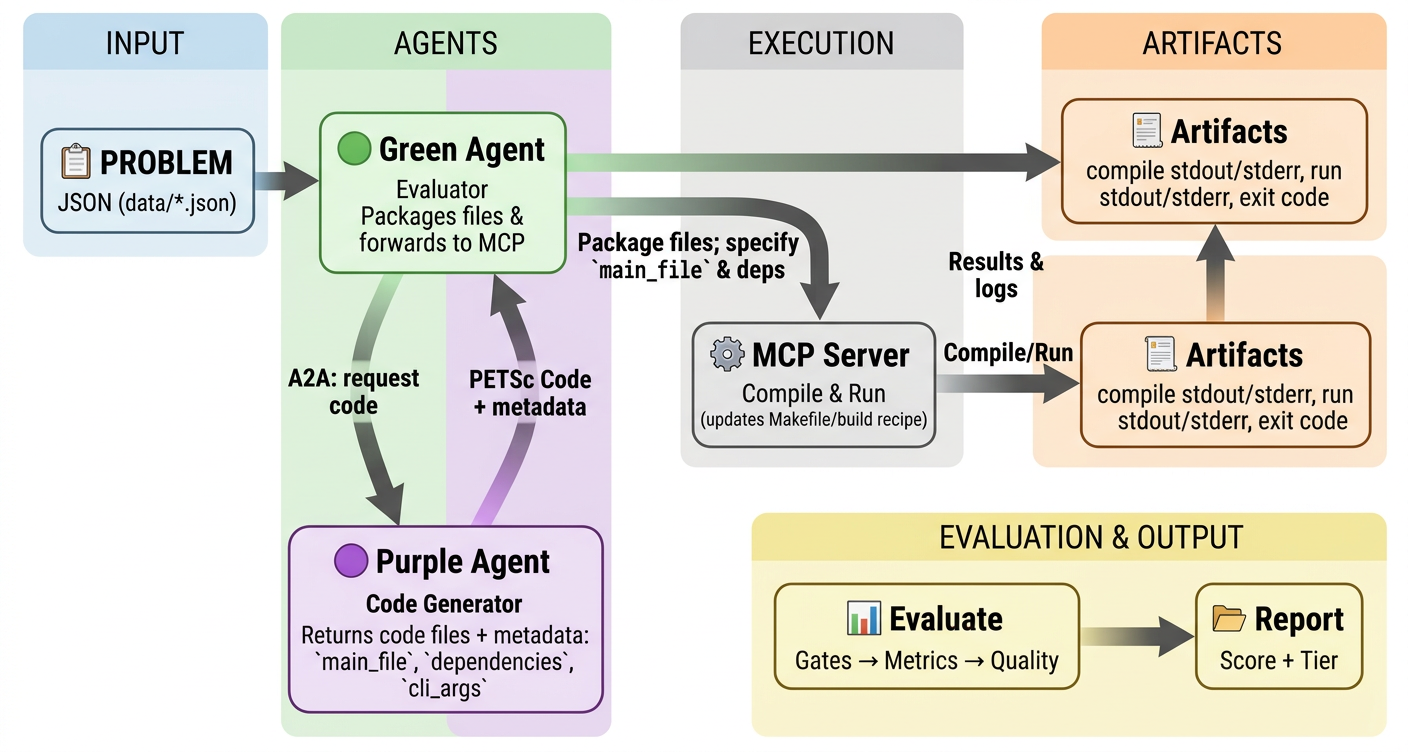}
  \caption{Overview of the \petscbench{} architecture.  The Green agent
    (evaluator) communicates with the Purple agent (model under test) through
    the A2A protocol and accesses compilation/execution tools through the MCP
    protocol.}
  \label{fig:architecture}
\end{figure*}

We combine these capabilities in an agents-evaluating-agents framework.
A tool-augmented evaluator agent compiles code against a library installation, executes it in a parallel environment, measures runtime performance, and applies both deterministic checks and LLM-based quality assessment.
We realize this idea in \petscbench{} (Fig.~\ref{fig:architecture}), choosing PETSc (with over
6{,}000 functions spanning time integrators, nonlinear solvers, linear solvers, data management, and optimization) as the test case, because
its breadth and complexity make the benchmark challenging and realistic.%
\footnote{\petscbench{} won third place in the Coding Agent track of the 2025-2026 AgentX-AgentBeats competition (\url{https://rdi.berkeley.edu/agentx-agentbeats.html}). The Green/Purple agent naming in this paper follows the AgentBeats convention.}

\newpage
Our contributions are:
\begin{enumerate}
\item \textbf{An agent-based evaluation framework for expert use of a
      production HPC library}, where a
      tool-augmented evaluator agent (Green agent) compiles, executes, and
      profiles code produced by a model-under-test agent (Purple agent) against
      a real HPC library installation, communicating through the
      Agent-to-Agent (A2A)~\cite{a2a2025} protocol
      and Model Context Protocol (MCP)~\cite{mcp2024} to enable black-box
      evaluation of interface-compatible coding agents without access to their internals.
\item \textbf{A multi-dimensional evaluation pipeline} with 14~evaluators
      across three stages (binary gates, quantitative metrics, and LLM-based
      quality assessments), aggregated into a confidence-weighted composite
      score.
\item \textbf{A curated PETSc benchmark suite} of six problems spanning
      key modules in PETSc and scientific domains.
\item \textbf{An empirical analysis} of coding agents backed by three frontier
      LLMs, revealing that current LLMs produce readable, well-structured code
      but struggle with correctness on challenging problems and with
      library-specific conventions even when the code compiles and runs, limitations that pass/fail benchmarks do not capture.
\end{enumerate}

\section{Related Work}
\label{sec:related}

LLMs are increasingly being used in scientific software engineering,
including code generation~\cite{chen2024ompgpt,nichols2024hpc,valerolara2023comparingllama2gpt3llms}, HPC software development~\cite{godoy2024large}, and compiler validation~\cite{munley2024llm4vv}.
Consequently, a growing body of benchmarks has emerged to evaluate LLM performance on programming and scientific computing tasks.

\medskip
\noindent\textbf{Code generation benchmarks.}
HumanEval~\cite{chen2021evaluating} and MBPP~\cite{austin2021program}
evaluate LLM-generated Python functions using unit tests, while
SWE-bench~\cite{jimenez2024swebench} and
CodeContests~\cite{li2022competition} extend evaluation to repository-level tasks
and algorithmic reasoning.
ScienceAgentBench~\cite{chen2025scienceagentbench},
AInsteinBench~\cite{duston2025ainsteinbench}, 
SciCode~\cite{tian2024scicode}, MMSciCode~\cite{xia2026mmscicode}, and
SciCodePile~\cite{sun2026scicodepile} further benchmark models and agents on
scientific coding tasks curated from research problems.
These benchmarks primarily assess task completion or functional correctness. Most do not evaluate whether generated code uses an external scientific library idiomatically, selects suitable numerical algorithms, or follows library-specific parallel programming conventions.

\medskip
\noindent\textbf{HPC and scientific code generation.}
ParEval~\cite{nichols2024pareval} benchmarks parallel code generation across
six programming models and has been extended to repository-level
tasks~\cite{davis2025parevalrepo}.
HPC-Coder-V2~\cite{chaturvedi2024hpccoder} fine-tunes LLMs for parallel programming
languages, while ChatHPC~\cite{valerolara2025chathpc} presents foundations for a productive and trustworthy AI-assisted
HPC ecosystem. Other work evaluates models such as
DeepSeek~\cite{nader2025deepseek} and studies performance
optimization~\cite{cui2025perfopt} and legacy code
translation~\cite{gupta2025kokkos,dearing2025lassi}.
Domain-specific benchmarks include
FEM-Bench~\cite{mohammadzadeh2025fembench} for finite elements,
CFDLLMBench~\cite{somasekharan2025cfdllmbench} for computational fluid
dynamics, and SciML Agents~\cite{gaonkar2025sciml} for ODE solver generation.
These efforts broaden evaluation to compiled and parallel languages, scientific problems, and, in some cases, numerical accuracy or performance. 
They nevertheless focus primarily on whether a generated implementation produces the expected result rather than whether it uses the abstractions and conventions of a mature numerical library as an expert would.

PETSc's continuous-integration harness appropriately verifies examples by matching program output against stored reference files.
It is designed to detect regressions, however, rather than to assess whether independently generated code selects appropriate solvers. 
\petscbench{} addresses this complementary evaluation problem by testing not only whether generated code produces the correct result, but also whether it uses PETSc solvers, data structures, and programming conventions appropriately.

\medskip
\noindent\textbf{Agent architectures and evaluation.}
LLM-as-a-judge~\cite{zheng2024judging} uses a language model to score model
outputs, and a growing body of software-engineering work adopts this paradigm to assess qualities that reference-based metrics miss, such as readability and adherence to best practices~\cite{he2026llmjudge}.
Although many such methods operate primarily on
textual representations, recent agent-based evaluators augment LLM
judgment with execution and other external evidence.
CodeVisionary~\cite{wang2025codevisionary} 
decomposes task requirements, gathers contextual evidence through tools such as dynamic execution, unit testing, and static analysis, and uses multiple judges to produce fine-grained assessments.
LLM4VV~\cite{munley2024llm4vv} similarly incorporates compilation and
execution evidence into an agent-based validation pipeline for
LLM-generated compiler tests.
ProjDevBench~\cite{lu2026projdevbench} combines
end-to-end execution with rule-based and LLM-assisted code review to
evaluate complete software projects.

Multi-agent frameworks such as
MOSAIC~\cite{raghavan2025mosaic}, PDE-Agent~\cite{liu2025pdeagent}, and
MooseAgent~\cite{zhang2025mooseagent} use agents for code \emph{generation}
but evaluate the results with traditional test suites.
Closest to our setting, our prior work developed retrieval-augmented coding
assistants for PETSc~\cite{smith2025petsc} and agent-based workflows to improve PETSc usability and productivity~\cite{smith2026improving}. These efforts focus on assisting PETSc use and development 
rather than at evaluating the quality of AI-generated PETSc code.
A related concern is benchmark rigor, since agentic benchmarks can leak solutions or reward gaming instead of capability~\cite{zhu2025agenticbenchmarks}.

Relative to these efforts, 
\petscbench{} targets a distinct evaluation setting: expert use of a production HPC library.
Its evaluator is itself a tool-augmented agent that compiles, executes, measures, and judges code through a multi-stage pipeline, communicating with the model under test via standardized protocols (A2A and MCP).
The benchmark also reduces opportunities for gaming and contamination by separating problem specifications from reference-output checkers, reporting per-category scores, and altering tutorial-like problems.
To our knowledge, \petscbench{} is the first agent-based benchmark designed specifically to evaluate whether AI-generated scientific HPC code uses a production numerical library in an expert, idiomatic, and parallel-aware manner.

\section{An Agentic Evaluation Framework for HPC Libraries}
\label{sec:architecture}

Figure~\ref{fig:architecture} shows the framework's three components.
The Green agent orchestrates evaluation, the Purple agent under test generates code, and the MCP server compiles and runs it against PETSc in a sandbox.

\subsection{Design Principles}
Evaluating library-based HPC code makes the evaluation infrastructure itself a substantial engineering problem.
It must manage a PETSc installation, compile against it, run binaries under MPI, recover from compilation and runtime failures, and coordinate a sequence of evaluators.
Bundling this into a single static harness would make the evaluator brittle and hard to extend.

An agentified architecture addresses this challenge by decoupling the model
under test and the evaluator into independent agents and making the evaluator an autonomous agent, not a fixed test harness.
In particular, our \emph{agents-evaluating-agents} architecture stands on three principles:
(1) role-based separation, where the evaluator and the model under
test are independent agents that interact solely through A2A, and any agent
speaking this protocol can be easily plugged into the evaluation
infrastructure;
(2) sandboxed execution, where generated code is compiled and run in a
sandboxed PETSc/MPI environment through MCP tools; and
(3) modular extensibility, where adding a benchmark problem requires
only a JSON specification and adding an evaluator requires implementing a single Python class.

\subsection{Component Overview}

The system consists of three major components as shown in Fig.~\ref{fig:architecture}:

\begin{itemize}
\item \textbf{Green agent (evaluator).}
      The Green agent is an assessment manager that orchestrates the benchmark. It loads problem specifications from JSON files, dispatches each problem to a Purple agent through A2A messages, receives generated code and auxiliary information, invokes MCP tools for compilation and execution, runs the 14-evaluator pipeline
      (see Sec.~\ref{sec:pipeline}), and prepares structured reports for artifact evaluation and leaderboard-style comparison.

\item \textbf{Purple agent (model under test).}
      The Purple agent is an LLM-based code-generation agent that receives a natural-language problem description through A2A and returns one or more C source files together with metadata including the entry-point file, dependencies, and command-line arguments.

\item \textbf{MCP server (tool provider).}
      The server provides a sandboxed execution environment exposing compilation and execution tools to the Green agent through MCP.
      It manages a custom PETSc installation, updates build
      recipes, compiles uploaded source files, runs the
      resulting binaries with MPI, and returns outputs (including errors and exit codes).
\end{itemize}

\subsection{Protocol Usage}

The framework relies on standardized protocols rather than a bespoke interface, so that any A2A-compatible agent can be evaluated as a black box and the growing ecosystem of MCP tools can be reused.
The interaction between the Green agent and Purple agent is governed by the
A2A protocol~\cite{a2a2025}.
The Green agent creates a \emph{task} containing the problem
description and the Purple agent responds with \emph{artifacts}, such as source files and metadata.
This decoupled setup allows the two agents to run on different machines, to sit behind firewalls, or to be hosted as cloud services.

The MCP~\cite{mcp2024} protocol governs the Green agent's
access to the compilation and execution environment.
The MCP server exposes two primary tools: (1) \texttt{compile}, which builds the uploaded code against PETSc,
and (2) \texttt{run}, which executes the binary with specified MPI ranks and
command-line arguments.
The reference implementation and public service endpoint are available at \url{https://github.com/petsc/petscagent-bench}.

\section{Evaluation Pipeline}
\label{sec:pipeline}

\begin{table*}[!t]
\centering
\caption{The 14 evaluators in the \petscbench{} pipeline.}
\label{tab:evaluators}
\begin{tabular}{lllllc}
\toprule
\textbf{\#} & \textbf{Evaluator} & \textbf{Stage} & \textbf{Category}
  & \textbf{Method} & \textbf{Conf.} \\
\midrule
\rowcolor{catCorrectness}
1  & Compilation         & Gate    & Correctness (syntactic)    & Deterministic   & 1.0 \\
\rowcolor{catCorrectness}
2  & Execution           & Gate    & Correctness (functional)    & Deterministic   & 1.0 \\
\rowcolor{catCorrectness}
3  & Memory Safety       & Gate    & Correctness (semantic)    & Runtime& 0.7--1.0\rlap{$^*$} \\
\rowcolor{catPETSc}
4  & API Usage           & Gate    & Library-specific          & Deterministic   & 1.0 \\
\midrule
\rowcolor{catCorrectness}
5  & Numerical Accuracy  & Metric  & Correctness (numerical)    & Runtime   & 1.0 \\
\rowcolor{catPerformance}
6  & Execution Time      & Metric  & Performance    & Runtime   & 1.0 \\
\midrule
\rowcolor{catCode}
7  & Readability         & Quality & Code    & LLM             & LLM \\
\rowcolor{catCode}
8  & Code Style          & Quality & Code    & LLM + static    & LLM \\
\rowcolor{catCode}
9  & Documentation       & Quality & Code    & LLM             & LLM \\
\rowcolor{catAlgorithm}
10 & Algorithm Approp.   & Quality & Appropriateness       & LLM             & LLM \\
\rowcolor{catAlgorithm}
11 & Solver Choice       & Quality & Appropriateness       & LLM + heuristic & LLM \\
\rowcolor{catPETSc}
12 & Best Practices      & Quality & Library-specific          & LLM + patterns  & LLM \\
\rowcolor{catPETSc}
13 & Error Handling      & Quality & Library-specific          & Deterministic   & 1.0 \\
\rowcolor{catPETSc}
14 & Parallel Awareness  & Quality & Library-specific          & Deterministic   & 0.8 \\
\bottomrule
\multicolumn{6}{@{}l}{\footnotesize $^*$1.0 with Valgrind; 0.7 without.}
\end{tabular}
\end{table*}

The pipeline operationalizes the principle from Section~\ref{sec:intro} that for library-based code, passing tests is necessary but not sufficient.
It is organized along two independent axes.
The first axis is the \emph{stage}, which controls execution order and cost.
Evaluation proceeds through three sequential stages, \emph{gates}, \emph{metrics}, and \emph{quality assessments}, in increasing order of expense.
Gates are hard filters.
If any gate fails, evaluation terminates with a zero score, so only code that compiles and runs reaches the more costly later stages.
This enforces the necessary baseline before the pipeline spends effort measuring the dimensions that pass/fail benchmarks overlook.

The second axis is the \emph{scoring category}, which determines how each evaluator contributes to the final score.
Every evaluator belongs to one of five categories: \textbf{correctness} (syntactic, functional, semantic, and numerical),
\textbf{performance} (execution time),
\textbf{code} (readability, style, documentation),
\textbf{appropriateness} (algorithm and solver selection), and
\textbf{library-specific} (API conventions, error handling, parallel awareness).
Stage and category are orthogonal.
A single category can draw on evaluators from several stages, as correctness does through the compilation gate, the numerical-accuracy metric, and semantic checks.
Table~\ref{tab:evaluators} lists all 14 evaluators with their stage, category, method, and confidence.

\subsection{Stage 1: Gates}

The four gates establish this necessary baseline. Evaluation continues only if the generated code passes all of them.
\begin{itemize}
\item \textbf{Compilation}. The C/CUDA/Kokkos source files compile against PETSc using PETSc's Makefile system.
\item \textbf{Execution}. The compiled binary runs to completion without crashes or errors.
\item \textbf{Memory safety.} The submitted code must have no leaks or invalid memory accesses. This gate leverages runtime analysis tools (e.g., Valgrind) or PETSc's own memory tracking features to detect memory issues.
\item \textbf{API usage.} The code must follow correct PETSc patterns such as
proper bracketing of \texttt{PetscInitialize} /\texttt{PetscFinalize} and using PETSc public header files.
\end{itemize}

\subsection{Stage 2: Metrics}
\label{sec:metrics}

Each metric evaluator produces a numerical score in $[0,1]$.

\medskip
\noindent\textbf{Numerical accuracy.}
The numerical output is compared against reference values from the problem specification.
We compute the relative error $\varepsilon = \|y_{\text{output}} - y_{\text{ref}}\| / \|y_{\text{ref}}\|$ and map it to a score with an exponential decay,
\begin{equation}
  s_{\text{acc}} = \min\!\bigl(1,\; e^{-\varepsilon / \tau}\bigr),
  \label{eq:accuracy}
\end{equation}
where $\tau$ is a configurable tolerance defaulting to $10^{-3}$.

\medskip
\noindent\textbf{Execution time.}
Runtime is scored by piecewise linear interpolation across four thresholds. The reported experiments use $t_1\!=\!0.5$\,s, $t_2\!=\!1$\,s, $t_3\!=\!2$\,s, and $t_4\!=\!4$\,s. Each problem may specify thresholds suited to its expected cost and target hardware.
The score is 1.0 for $t \le t_1$ and decays linearly through 0.8, 0.6, and 0.2 at $t_2$, $t_3$, and $t_4$.
For $t > t_4$ it falls off as $0.2\,t_4/t$.
The current performance metric measures execution time for these benchmark problems on a fixed platform; it is not intended to measure parallel scalability.

\subsection{Stage 3: Quality Assessments}

The final and most expensive stage applies eight evaluators (Table~\ref{tab:evaluators}), each producing a score in $[0,1]$, across the \textbf{code}, \textbf{appropriateness}, and \textbf{library-specific} categories.
Six of the eight rely on LLM judgment, optionally augmented by static analysis or heuristics.
The other two are library-specific and use deterministic checks.
We reserve LLM-based evaluation for the subjective or semantic dimensions that deterministic techniques cannot capture, such as whether code is readable or whether a solver choice is appropriate, and use deterministic checks wherever a dimension can be verified directly.
Evaluators without data dependencies run concurrently within a stage to reduce latency.

Because LLM-based evaluators can be sensitive to prompts and model versions,
their scores should be interpreted as structured expert-assistance signals, not ground truth.
We make each evaluator emit a rationale and confidence value, and the final report preserves the per-category scores so users can inspect whether a composite score is driven by deterministic failures, numerical accuracy, or subjective quality judgments.
In the current benchmark release, we therefore use composite scores primarily for ranking and diagnosis within a fixed evaluator configuration.
Validating LLM-based quality scores against independent PETSc expert review is an important next step.

\subsection{Score Aggregation}
\label{sec:aggregation}

Individual evaluator scores are aggregated in two steps to produce the final composite score in $[0,100]$.

\medskip
\noindent\textbf{Step 1: Category scores.}
Each evaluator $j$ in category $c$ gives a score $s_j$ in $[0,1]$ with confidence
$\alpha_j$.
The scores are weighted by confidence and aggregated into a single category score $S_c$ for each of the five categories:
\begin{equation}
  S_c = \frac{\sum_{j \in c} \alpha_j \, s_j}{\sum_{j \in c} \alpha_j}.
  \label{eq:category}
\end{equation}
Confidence is a variable returned by the LLM for LLM-based evaluators and a fixed value for deterministic evaluators.

\medskip
\noindent\textbf{Step 2: Composite score.}
Scores across the categories are rescaled to 0--100 and then combined with configurable
weights $w_c$ to produce the final composite score:
\begin{equation}
  S_{\text{composite}} = \sum_{c} w_c \, S_c.
  \label{eq:composite}
\end{equation}
The default weights are
{\setlength{\fboxsep}{1.5pt}%
\colorbox{catCorrectness}{correctness} 0.35,
\colorbox{catPerformance}{performance} 0.15,
\colorbox{catCode}{code} 0.15,
\colorbox{catPETSc}{library-specific} 0.20,
\colorbox{catAlgorithm}{appropriateness} 0.15.}
Correctness and library-specific quality receive the largest weights because failures in either dimension can invalidate production scientific code.
The weights define the benchmark policy and remain configurable for other applications.
The composite score is not intended to replace the category breakdown.
For example, a submission can obtain a moderate score by passing gates and running quickly while still failing PETSc-specific best practices.
Conversely, readable code can score poorly if numerical accuracy or execution fails.
We therefore report category-level scores and gate failures alongside the composite score in all analyses.

\section{Benchmark Suite}
\label{sec:benchmark}

\begin{table*}[t]
\centering
\caption{Benchmark problems in \petscbench{}. Module indicates the primary
  PETSc component exercised; difficulty is assessed on a three-level scale.}
\label{tab:problems}
\begin{tabular}{@{}lllll@{}}
\toprule
\textbf{Problem} & \textbf{Module} & \textbf{Difficulty} & \textbf{Domain}
  & \textbf{Key Challenge} \\
\midrule
Robertson ODE
  & TS & Medium & Stiff ODE system
  & Stiffness, adaptive stepping \\
1D Advection
  & TS & Easy & Hyperbolic PDE
  & Upwind discretization, periodic BCs \\
Rosenbrock Opt.
  & Tao & Easy & Nonlinear optimization
  & Gradient computation, convergence \\
Darcy Flow
  & KSP, DM & Hard & Elliptic PDE (FEM)
  & FE assembly, heterog.\ coefficients \\
2D Navier--Stokes
  & TS, SNES & Hard & Incompressible flow
  & Staggered grid, implicit coupling \\
Vec/MPI Scatter
  & Vec & Easy & Data management
  & MPI communication, VecScatter \\
\bottomrule
\end{tabular}
\end{table*}

We curate a suite of six problems (Table~\ref{tab:problems}) designed to probe whether coding agents can use PETSc as an expert would.
The selection balances several goals.
The problems exercise distinct PETSc subsystems, spanning time integration (TS), optimization (Tao), linear solvers and mesh management (KSP, DM), and distributed data (Vec), and they cover a range of numerical methods from ODE integration through hyperbolic and elliptic PDEs to nonlinear optimization and parallel data management, each demanding a different mode of reasoning.
They also span a difficulty range from easy to hard, so the suite discriminates among models rather than saturating or defeating all of them.
Where a problem resembles an existing PETSc tutorial, we modify it so that memorized solutions do not transfer, as detailed at the end of this section.
We intend these six problems as an initial benchmark and expect the suite to expand across more of PETSc's functionality.

Each problem is specified as a JSON file with a natural-language description of the governing equations, boundary conditions, and requirements, together with test cases giving command-line arguments, MPI rank counts, time limits, and reference outputs.
The descriptions are deliberately underspecified.
We withhold problem characteristics that an expert would have to infer, such as the stiffness of an ODE, the appropriate solver, or the specific PETSc API, so that success requires reasoning about the problem instead of transcribing a known recipe.
This design lets the suite test library-selection skill rather than recall.

The problems fall into three difficulty tiers that probe increasing levels of this skill.
The easy problems, Rosenbrock optimization, 1D advection, and Vec/MPI scatter, test whether an agent can wire up the correct PETSc interface for a well-posed task, including Tao callback signatures, upwind discretization with periodic boundaries, and distributed vector communication through VecScatter.
The medium problem, Robertson ODE, further requires recognizing a numerical property, stiffness, and responding with an implicit integrator and appropriate tolerances.
The hard problems, Darcy flow and 2D Navier--Stokes, demand composing multiple subsystems, finite-element assembly over heterogeneous coefficients for the former and coupled multi-field nonlinear solves on a staggered grid for the latter.
Reference outputs come from hand-written PETSc implementations and are checked with problem-specific tolerances, while the API and quality evaluators judge whether the generated code follows PETSc conventions beyond matching the final number.

PETSc's source repository is public and almost certainly part of LLM training data.
To mitigate data
contamination~\cite{jacovi2023stop},
we intentionally alter initial conditions, boundary conditions, or discretizations if similar problems exist in PETSc tutorials.
These changes reduce the likelihood that models can simply reproduce memorized tutorial solutions, but they do not eliminate all contamination risk.
For example, PETSc's tutorial suite solves Darcy flow with
finite differences.
Our specification requires a finite-element
formulation instead, forcing the model to use higher-level
abstractions such as PETSc's \texttt{DMPLEX} module for mesh management and element assembly.

\section{Evaluation Results and Analysis}
\label{sec:results}

We evaluate three frontier coding agents on the benchmark.
Across the board, the models produce readable, well-structured code yet struggle with correctness on hard problems and with PETSc-specific conventions throughout, dimensions that pass/fail benchmarks would not surface.
The following analysis breaks this pattern down by problem, by category, and by failure mode.


\begin{figure}[!tb]
\centering
\includegraphics[width=\linewidth]{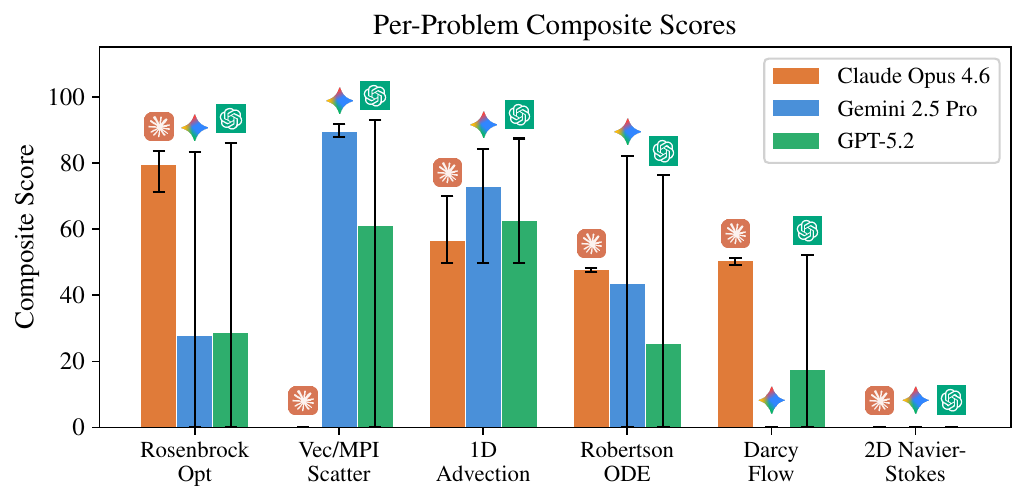}
\caption{Per-problem composite scores. Bars show the mean over three runs
  per model, counting gate failures as zero; error bars span the minimum
  and maximum of those runs.}
\label{fig:perproblem}
\end{figure}

\subsection{Evaluation Settings}

Each agent uses a frontier LLM (Claude Opus 4.6, Gemini 2.5 Pro, or GPT-5.2) and is given only one shot to generate code to solve each problem.
Each test is repeated three times under identical conditions to account for variability in LLM outputs.
The Green agent uses a single fixed model, GPT-5.2, for all LLM-based quality evaluators, so that every submission is judged under identical criteria and quality scores remain comparable across models.
All models are run at temperature~0 to aid reproducibility, although hosted APIs may still introduce implementation-level nondeterminism.
We also pin the completion-token limit to 32{,}000, because the API default of 4{,}096 truncates and thus invalidates the longest solutions.
Each Purple agent is evaluated in a one-shot setting.
It receives the natural-language specification once and gets no compiler errors or test feedback for iterative repair.
The protocol measures unaided first-attempt performance from the problem specification alone.
Retrieval, planning, and repair can be evaluated as separate agent configurations.
The Green agent compiles each submission against a fixed PETSc/MPI environment. 
The experiments used PETSc 3.25.3 with GCC 11.4.0 and MPICH 5.0.0 on an 8-core Intel Core i7-11700KF workstation with 32 GB RAM; complete environment details are provided in the artifact appendix.
The Green agent records source files, compilation commands, stdout/stderr, exit codes, metric outputs, evaluator rationales, and final scores.
The artifact package (available at \url{https://github.com/petsc/petscagent-bench}) contains the problem JSON files, reference-output checkers, evaluator prompts and rubrics, configuration files, and scripts required to replay cached submissions and rerun the benchmark with new A2A-compatible agents.

\subsection{Overall Performance}

All three models achieve only modest average composite scores.
Claude Opus~4.6 and Gemini~2.5~Pro are effectively tied at \textbf{39.0} and \textbf{38.9}, with GPT-5.2 trailing at \textbf{32.6}.
Of the 54~total attempts, 25 score zero from gate failures while 13 exceed~80, reflecting a benchmark that separates clear successes from clear failures.
These averages summarize the 54 observed attempts: Claude and Gemini differ by 0.1 points, and GPT-5.2 trails them by 6.3--6.4 points.
All three models become less reliable on problems that require both numerical reasoning and detailed PETSc knowledge.

Although GPT-5.2 is both a model under test and the quality judge, we do not observe an obvious self-preference effect in the aggregate results.
The judge scores only the subjective quality dimensions, which carry a minority of the composite weight, while the heaviest category, correctness at 0.35, and performance are measured deterministically.
Using the same judge for every submission keeps the rubric fixed across models, but it does not eliminate the possibility of model- or style-specific judge bias.
GPT-5.2 places last overall despite judging its own code, which suggests that any such bias is not large enough to dominate the reported ranking; we nevertheless treat judge-model bias as a limitation.

\subsection{Per-Problem Analysis}

A clear difficulty gradient can be seen in Figure~\ref{fig:perproblem}.
When gates are passed, the easier problems score well, with Rosenbrock averaging 79--86 and Vec/MPI~Scatter 90--92, but neither is passed reliably by every model.
Claude fails all three Vec/MPI attempts, while Gemini and GPT-5.2 each fail Rosenbrock in two of three runs.
Medium problems exhibit greater variability.
1D~Advection averages 57--73 and passes every gate, while Robertson~ODE scores 48--76 when it runs but clears gates in only six of nine attempts.
The hard problems, Darcy~Flow and 2D~Navier--Stokes, defeat
nearly all attempts.
On Darcy~Flow, Claude passes all three runs at an average of 51 and GPT-5.2 passes once at 52, while Gemini fails every attempt.
No model passes 2D~Navier--Stokes.

All three models fail on the Navier--Stokes problem, but for distinct reasons.
Gemini~2.5~Pro fails at compilation in all three runs, passing
incorrect argument types to \texttt{DMStagCreate2d}, for example
supplying \texttt{NULL} where PETSc expects a \texttt{PetscInt}
stencil width, indicating a fundamental misunderstanding of the
DMStag API. 
Claude Opus~4.6 and GPT-5.2 both compile successfully but then crash at
runtime.
In five of the six runs, PETSc complains about
``Matrix is missing diagonal entries,'' indicating that the Jacobian
assembly omits some grid points.
One of GPT-5.2's attempts fails with
\texttt{DIVERGED\_NONLINEAR\_SOLVE}, implying a problematic nonlinear solver configuration.
None of the three frontier models solves this case in any of its three one-shot attempts.

\subsection{Gate Failure Analysis}

Gate pass rates are \textbf{67\%} for Claude, \textbf{50\%} for Gemini,
and \textbf{44\%} for GPT-5.2.
Among submissions that reach successful execution, all pass the remaining memory-safety and API-usage gates.
All 29 such submissions used the standard-error memory screen at confidence 0.7.
This locates the primary gate bottlenecks before the later quality stages: first in generating code that compiles against PETSc, and then in avoiding runtime failures on the hardest problems.
Gate reliability differs substantially across models, even though their composite scores are close.
For all three agents, however, every gate failure is a compilation or runtime failure.

\subsection{Category-Level Analysis}

\begin{figure}[!tb]
\centering
\includegraphics[width=0.78\linewidth]{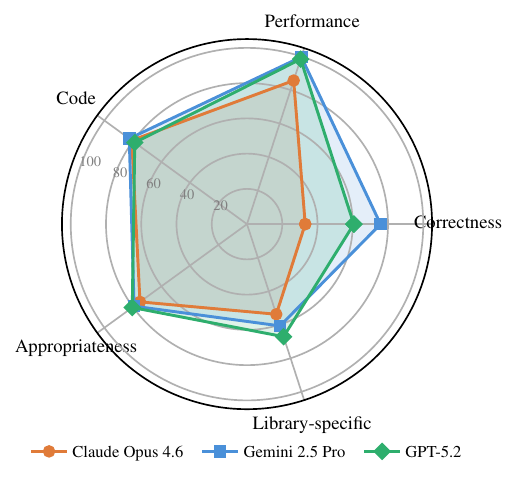}
\caption{Category-level scores on a 0--100 scale, averaged over
  gate-passed attempts only (Claude: $n=12$; Gemini: $n=9$; GPT-5.2:
  $n=8$, of 18 attempts each).}
\label{fig:categories}
\end{figure}

Figure~\ref{fig:categories} shows broadly similar profiles across models.
Scores are averaged over gate-passed attempts only.
\textbf{Performance} is uniformly high at 86--100, reflecting fast
code execution across all problems.
\textbf{Code}, 79--83, and \textbf{appropriateness}, 75--81, are
also strong, indicating that the agents tend to generate readable,
well-structured code with reasonable choices of numerical method and solvers.
\textbf{Correctness} varies widely by model and problem, from 33 to 76,
as some problems, particularly stiff ones, are highly sensitive to solver parameters and numerical stability.
\textbf{Library-specific} scores are uniformly weak, never exceeding 67 for any model, driven by systematic failures in PETSc-specific error handling and best-practice compliance.

\subsection{Qualitative Findings}

Our multidimensional evaluation highlights several informative cases.
For the Vec/MPI Scatter problem, Claude triggers segmentation faults in three attempts despite otherwise high-quality code.
On the stiff Robertson~ODE problem, Gemini~2.5~Pro produces clean, well-structured code but poor numerical accuracy due to loose integrator tolerances it selects.
Error-handling scores are especially low, exactly 0.30 in 20 of 29 gate-passed attempts, because most models do not wrap PETSc calls in \texttt{PetscCall()}.
This suggests that current LLMs have learned PETSc usage from older versions (pre-3.17) that used the legacy
\texttt{CHKERRQ()} style.

\subsection{Robustness to the Choice of Judge}

\begin{table}[!tb]
\centering
\caption{Two judges scoring the same generated programs. A dash marks a
  gate failure, scored zero by either judge; the composite covers all
  six problems. 2D Navier–Stokes is omitted because all three cached submissions fail a gate and therefore receive zero under either judge.}
\label{tab:judge}
\setlength{\tabcolsep}{2.6pt}
\setlength{\aboverulesep}{0pt}
\setlength{\belowrulesep}{0pt}
\setlength{\extrarowheight}{0.75ex}
\small
\begin{tabular}{@{}llccccc>{\columncolor{compositeTint}}c@{}}
\toprule
& & \multicolumn{5}{c}{\textbf{Test problem}} & \\
\cmidrule(lr){3-7}
\textbf{Model} & \textbf{Judge} & \textbf{Rosen.} & \textbf{Robert.}
  & \textbf{Vec} & \textbf{Advec.} & \textbf{Darcy} & \textbf{Comp.} \\
\midrule
\modellogo{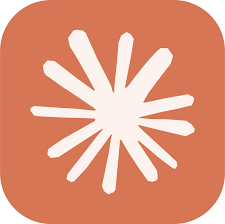}~Claude
  & GPT & 83.8 & 47.6 & --   & 70.1 & 51.4 & 42.1 \\
  & Claude & 84.7 & 49.0 & --   & 69.5 & 50.3 & 42.3 \\
\midrule
\modellogo{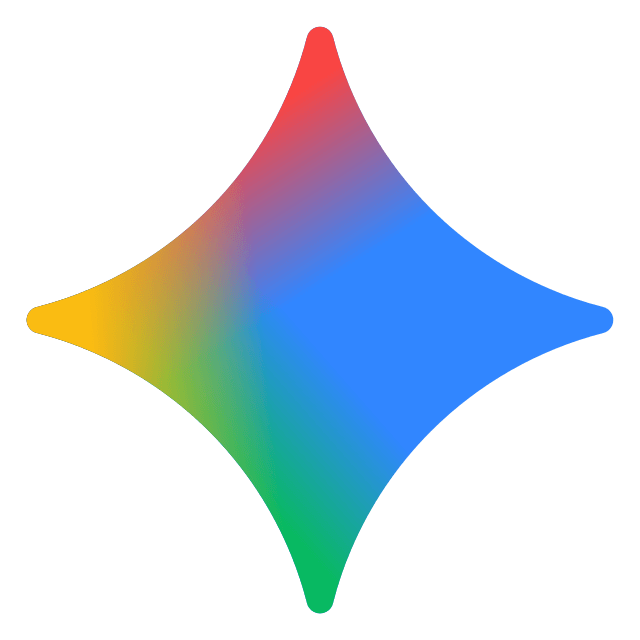}~Gemini
  & GPT & 87.6 & 49.3 & 87.9 & 49.7 & --   & 45.8 \\
  & Claude & 87.7 & 49.3 & 86.0 & 49.6 & --   & 45.4 \\
\midrule
\modellogo{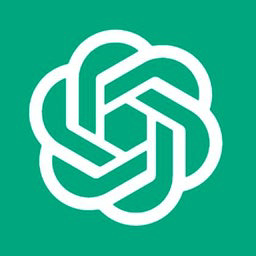}~GPT-5.2
  & GPT & 84.7 & 78.1 & 93.4 & 82.5 & --   & 56.4 \\
  & Claude & 86.6 & 77.2 & 88.9 & 82.2 & --   & 55.8 \\
\bottomrule
\end{tabular}
\end{table}

To test whether the quality scores depend on the choice of judge, we rerun the quality evaluators with a second judge, Claude Opus~4.6.
We cache the code generated in one run and replay it against the second judge, so that both judges see byte-identical programs.
They differ by \textbf{1.1} points on average over the 12~gate-passed attempts shown in Table~\ref{tab:judge}.
The largest disagreement is 4.5~points, and neither judge scores consistently higher.
The composite scores agree within \textbf{0.6} points for every model, far less than the several-point shift seen when the same model and judge are rerun on freshly generated code.
In this experiment, generation variability exceeds the score variation introduced by substituting the second judge.

\section{Discussion}
\label{sec:discussion}

\noindent\textbf{Why library-based evaluation matters.}
The dimensions that separate high from low scores, including error-handling
conventions, solver selection, and resource-management patterns, are
inherently library-specific and do not arise in from-scratch
implementations.
By targeting PETSc, \petscbench{} measures the skill that production HPC code actually demands, composing scalable components correctly instead of reimplementing known algorithms.

Our results show that models fail at this skill in two distinct ways, and the distinction matters because the two have very different remedies.
One failure mode appears to reflect knowledge staleness.
The widespread use of the deprecated \texttt{CHKERRQ()} macro instead of \texttt{PetscCall()} reflects training data that predates PETSc~3.17, suggesting a problem that retrieval of current documentation may mitigate. 
Failures in assembling Jacobians for finite-element methods and in matching integrator tolerances to problem stiffness reflect gaps in numerical reasoning, not stale facts, and these will not be closed by fresher documentation alone.

\medskip
\noindent\textbf{Implications for agent design.}
These two failure modes call for different interventions.
Knowledge-staleness failures are the more tractable.
Retrieval augmentation over current PETSc documentation could supply the idioms models are missing, and the prevalence of compilation failures suggests that multi-turn agents with compile-fix loops may improve pass rates, since compiler diagnostics are usually informative enough to guide corrections.
Numerical-reasoning failures are harder and unlikely to yield to context or retries alone.
The framework can measure gains from retrieval, planning, and repair under the same tasks and scoring criteria.

\medskip
\noindent\textbf{Limitations.}
The benchmark does not yet cover PETSc's full breadth, such as GPU kernels,
multiphysics, advanced preconditioners, performance portability, and large-scale distributed-memory scaling, but can be extended with additional JSON problem specifications and evaluator classes.
The present runtime metric compares fixed test instances on one workstation.
Multi-node scaling studies require parameterized problem sizes and rank counts.
In addition, LLM-based evaluators may vary across runs due to the inherent stochasticity of LLMs, though this is mitigated by temperature~0, fixed prompts, preserved rationales, and confidence weighting.
An expert-audit study could compare general-purpose and domain-adapted judges against independent PETSc maintainers on a fixed set of submissions.
The current results are also limited to one-shot generation.
Multi-turn agents that can read compiler errors and repair submissions may obtain different results.
Finally, modifying tutorial-like problems reduces but does not eliminate training-data contamination risk.

\medskip
\noindent\textbf{Responsible release and benchmark use.} 
The benchmark does not release models or human-subject data, but benchmark artifacts can still shape model-development incentives.
To reduce leaderboard gaming, the artifact separates public problem specifications from reference-output checkers and preserves per-category reports instead of only a single score.
The main intended use is diagnostic evaluation of scientific-code agents, not certification that generated code is safe for production scientific computing.
Users should still review generated code manually before using it in research or engineering workflows.

\section{Conclusion and Future Work}
\label{sec:conclusion}

We introduced \petscbench{}, a multidimensional benchmark and agent-based evaluation framework
for AI-generated scientific code that targets \emph{library-based}
HPC programming rather than from-scratch implementations.
Its evaluator is a tool-augmented agent that builds and runs each submission
against a real PETSc installation and scores it along multiple dimensions,
communicating with the model under test via standardized protocols.
This design supports black-box evaluation that goes beyond traditional
pass/fail benchmarking.

Our evaluation of Claude Opus~4.6, Gemini~2.5~Pro, and GPT-5.2 shows
that frontier LLMs generate readable, well-structured scientific code but
fall short on correctness and library-specific conventions, dimensions that traditional pass/fail evaluation does not capture.
Replaying identical programs past a second judge moves the quality scores by
about a point, 
suggesting that the main findings are not sensitive to the choice between these two judges.
We are extending the suite to GPU test problems and additional PETSc subsystems.
More broadly, as LLMs take on a growing share of library-based scientific programming, evaluating whether they use libraries the way an expert would becomes essential, and \petscbench{} is a step toward making that skill measurable.

\section*{Acknowledgments}
This material is based upon work partially supported by Laboratory Directed Research and Development (LDRD) funding from Argonne National Laboratory, provided by the Director, Office of Science, of the U.S. Department of Energy (DOE) under Contract No. DE-AC02-06CH11357. This work was also partially supported by the U.S. DOE, Office of Science, Office of Advanced Scientific Computing Research, by the Competitive Portfolios for Advanced Scientific Computing Research Program, by the Scientific Discovery through Advanced Computing (SciDAC) Program through the FASTMath Institute, and by the U.S. DOE Office of Science Distinguished Scientist Fellows Program.
This research used resources of the Argonne Leadership Computing Facility, a U.S. Department of Energy (DOE) Office of Science user facility at Argonne National Laboratory.

\bibliographystyle{IEEEtran}
\bibliography{references}

@techreport{petsc-user-ref,
  author      = {Satish Balay and Shrirang Abhyankar and Mark~F. Adams and others},
  title       = {{PETSc/TAO} Users Manual},
  institution = {Argonne National Laboratory},
  number      = {ANL-21/39 -- Revision 3.25},
  doi = {10.2172/2998643},
  url = {https://petsc.org/release/manual/manual.pdf},
  year        = {2026},
}

@article{chen2021evaluating,
  author  = {Mark Chen and Jerry Tworek and Heewoo Jun and others},
  title   = {Evaluating Large Language Models Trained on Code},
  journal = {arXiv preprint arXiv:2107.03374},
  year    = {2021},
}

@article{austin2021program,
  author  = {Jacob Austin and Augustus Odena and Maxwell Nye and others},
  title   = {Program Synthesis with Large Language Models},
  journal = {arXiv preprint arXiv:2108.07732},
  year    = {2021},
}

@inproceedings{jimenez2024swebench,
  author    = {Carlos~E. Jimenez and John Yang and Alexander Wettig and
               Shunyu Yao and Kexin Pei and Ofir Press and Karthik Narasimhan},
  title     = {{SWE-bench}: Can Language Models Resolve Real-World {GitHub} Issues?},
  booktitle = {Proceedings of the 12th International Conference on
               Learning Representations (ICLR)},
  year      = {2024},
}

@article{li2022competition,
  author  = {Yujia Li and David Choi and Junyoung Chung and others},
  title   = {Competition-Level Code Generation with {AlphaCode}},
  journal = {Science},
  volume  = {378},
  number  = {6624},
  pages   = {1092--1097},
  year    = {2022},
}

@inproceedings{zheng2024judging,
  author    = {Lianmin Zheng and Wei-Lin Chiang and Ying Sheng and others},
  title     = {Judging {LLM}-as-a-Judge with {MT-Bench} and {Chatbot Arena}},
  booktitle = {Advances in Neural Information Processing Systems (NeurIPS),
               Datasets and Benchmarks Track},
  volume    = {36},
  year      = {2023},
}

@misc{a2a2025,
  author       = {{Google}},
  title        = {Agent-to-Agent ({A2A}) Protocol Specification},
  year         = {2025},

}

@misc{mcp2024,
  author       = {{Anthropic}},
  title        = {Model Context Protocol ({MCP}) Specification},
  year         = {2024},

}

@inproceedings{smith2025petsc,
  author    = {Barry Smith and Junchao Zhang and Hong Zhang and others},
  title     = {{AI} Assistants to Enhance and Exploit the {PETSc}
               Knowledge Base},
  booktitle = {54th International Conference on Parallel Processing
               Companion (ICPP)},
  publisher = {ACM},
  note = {\doi{10.1145/3750720.3757281}},
  year      = {2025},
}

@article{mohammadzadeh2025fembench,
  author  = {Saeed Mohammadzadeh and Erfan Hamdi and Joel Shor and
             Emma Lejeune},
  title   = {{FEM-Bench}: A Structured Scientific Reasoning Benchmark
             for Evaluating Code-Generating {LLMs}},
  journal = {arXiv preprint arXiv:2512.20732},
  year    = {2025},
}

@article{duston2025ainsteinbench,
  author  = {Titouan Duston and Shuo Xin and Yang Sun and
             Daoguang Zan and others},
  title   = {{AInsteinBench}: Benchmarking Coding Agents on Scientific
             Repositories},
  journal = {arXiv preprint arXiv:2512.21373},
  year    = {2025},
}

@article{somasekharan2025cfdllmbench,
  author  = {Nithin Somasekharan and Ling Yue and Yadi Cao and
             Weichao Li and Patrick Emami and others},
  title   = {{CFDLLMBench}: A Benchmark Suite for Evaluating Large
             Language Models in Computational Fluid Dynamics},
  journal = {arXiv preprint arXiv:2509.20374},
  year    = {2025},
}

@article{gaonkar2025sciml,
  author  = {Saarth Gaonkar and Xiang Zheng and Haocheng Xi and
             Rishabh Tiwari and Kurt Keutzer and Dmitriy Morozov and
             Michael~W. Mahoney and Amir Gholami},
  title   = {{SciML} Agents: Write the Solver, Not the Solution},
  journal = {arXiv preprint arXiv:2509.09936},
  year    = {2025},
}

@article{liu2025pdeagent,
  author  = {Jianming Liu and Ren Zhu and Jian Xu and Kun Ding and
             Xu-Yao Zhang and Gaofeng Meng and Cheng-Lin Liu},
  title   = {{PDE-Agent}: A Toolchain-Augmented Multi-Agent Framework
             for {PDE} Solving},
  journal = {arXiv preprint arXiv:2512.16214},
  year    = {2025},
}

@inproceedings{chen2025scienceagentbench,
  author    = {Ziru Chen and Shijie Chen and Yuting Ning and
               Qianheng Zhang and others},
  title     = {{ScienceAgentBench}: Toward Rigorous Assessment of
               Language Agents for Data-Driven Scientific Discovery},
  booktitle = {Proceedings of the 13th International Conference on
               Learning Representations (ICLR)},
  year      = {2025},
}

@article{zhang2025mooseagent,
  author  = {Tao Zhang and Zhenhai Liu and Yong Xin and
             Yongjun Jiao},
  title   = {{MooseAgent}: A {LLM} Based Multi-Agent Framework for
             Automating {MOOSE} Simulation},
  journal = {arXiv preprint arXiv:2504.08621},
  year    = {2025},
}

@inproceedings{nichols2024pareval,
  author    = {Daniel Nichols and Joshua~H. Davis and Zhaojun Xie and
               Arjun Rajaram and Abhinav Bhatele},
  title     = {Can Large Language Models Write Parallel Code?},
  booktitle = {Proceedings of the 33rd International Symposium on
               High-Performance Parallel and Distributed Computing
               (HPDC)},
  year      = {2024},
}

@inproceedings{chen2025beyond,
  title={Beyond code pairs: Dialogue-based data generation for llm code translation},
  author={Chen, Le and Xu, Nuo and Chen, Winson and Lei, Bin and Lin, Pei-Hung and Zhou, Dunzhi and Thakur, Rajeev and Ding, Caiwen and Jannesari, Ali and Liao, Chunhua},
  booktitle={Proceedings of the 64th Annual Meeting of the Association for Computational Linguistics (Volume 1: Long Papers)},
  pages={33781--33803},
  year={2026}
}

@inproceedings{davis2025parevalrepo,
  author    = {Joshua~H. Davis and Daniel Nichols and Ishan Khillan and
               Abhinav Bhatele},
  title     = {{ParEval-Repo}: A Benchmark Suite for Evaluating {LLMs}
               with Repository-level {HPC} Translation Tasks},
  booktitle = {Proceedings of the 54th International Conference on
               Parallel Processing (ICPP)},
  pages     = {94--103},
  year      = {2025},
  doi       = {10.1145/3754598.3754669},
}

@inproceedings{chaturvedi2024hpccoder,
  author    = {Aman Chaturvedi and Daniel Nichols and Siddharth Singh and
               Abhinav Bhatele},
  title     = {{HPC-Coder-V2}: Studying Code {LLMs} Across Low-Resource
               Parallel Languages},
  booktitle = {Proceedings of the International Conference on
               High Performance Computing (ISC)},
  pages     = {1--14},
  year      = {2025},
}

@inproceedings{nader2025deepseek,
  author    = {Noujoud Nader and Patrick Diehl and Steve Brandt and
               Hartmut Kaiser},
  title     = {{LLM \& HPC}: Benchmarking {DeepSeek}'s Performance in
               High-Performance Computing Tasks},
  booktitle = {Proceedings of the International Conference on
               High Performance Computing (ISC)},
  pages     = {626--638},
  year      = {2025},
}

@article{cui2025perfopt,
  author  = {Bowen Cui and Tejas Ramesh and Oscar Hernandez and
             Keren Zhou},
  title   = {Do Large Language Models Understand Performance
             Optimization?},
  journal = {arXiv preprint arXiv:2503.13772},
  year    = {2025},
}

@article{gupta2025kokkos,
  author  = {Sparsh Gupta and Kamalavasan Kamalakkannan and
             Maxim Moraru and Galen Shipman and Patrick Diehl},
  title   = {From Legacy {Fortran} to Portable {Kokkos}: An Autonomous
             Agentic {AI} Workflow},
  journal = {arXiv:2509.12443},
  year    = {2025},
}

@article{dearing2025lassi,
  author  = {Matthew~T. Dearing and Yiheng Tao and Xingfu Wu and
             Zhiling Lan and Valerie Taylor},
  title   = {Leveraging {LLMs} to Automate Energy-Aware Refactoring
             of Parallel Scientific Codes},
  journal = {arXiv preprint arXiv:2505.02184},
  year    = {2025},
}

@article{raghavan2025mosaic,
  author  = {Siddeshwar Raghavan and Tanwi Mallick},
  title   = {{MOSAIC}: Multi-agent Orchestration for Task-Intelligent
             Scientific Coding},
  journal = {arXiv preprint arXiv:2510.08804},
  year    = {2025},
}

@inproceedings{jacovi2023stop,
  author    = {Alon Jacovi and Avi Caciularu and Omer Goldman and Yoav Goldberg},
  title     = {Stop Uploading Test Data in Plain Text: Practical Strategies for
               Mitigating Data Contamination by Evaluation Benchmarks},
  booktitle = {Proceedings of the 2023 Conference on Empirical Methods in
               Natural Language Processing (EMNLP)},
  year      = {2023},
}

@book
{
  greenshields2025,
  title     = "OpenFOAM v13 User Guide",
  author    = "Greenshields, Christopher",
  year      = 2025,
  url       = "https://doc.cfd.direct/openfoam/user-guide-v13",
  publisher = "The OpenFOAM Foundation",
  address   = "London, UK"
}

@inproceedings{valerolara2025chathpc,
  author    = {Pedro Valero-Lara and Aaron R. Young and Jeffrey S. Vetter and
               Zheming Jin and Swaroop Pophale and Mohammad Alaul Haque Monil and
               Keita Teranishi and William F. Godoy},
  title     = {{ChatHPC}: Building the Foundations for a Productive and
               Trustworthy {AI}-Assisted {HPC} Ecosystem},
  booktitle = {Proceedings of the International Conference for High Performance
               Computing, Networking, Storage and Analysis (SC)},
  year      = {2025},
  doi       = {10.1145/3712285.3759787},
}

@inproceedings{chen2024ompgpt,
  title={{OMPGPT}: A generative pre-trained transformer model for {OpenMP}},
  author={Chen, Le and Bhattacharjee, Arijit and Ahmed, Nesreen and Hasabnis, Niranjan and Oren, Gal and Vo, Vy and Jannesari, Ali},
  booktitle={European Conference on Parallel Processing},
  pages={121--134},
  year={2024},
  organization={Springer}
}

@inproceedings{nichols2024hpc,
  title={{HPC-Coder}: Modeling parallel programs using large language models},
  author={Nichols, Daniel and Marathe, Aniruddha and Menon, Harshitha and Gamblin, Todd and Bhatele, Abhinav},
  booktitle={ISC High Performance 2024 Research Paper Proceedings (39th International Conference)},
  pages={1--12},
  year={2024},
  organization={Prometeus GmbH}
}

@article{godoy2024large,
  title={Large language model evaluation for high-performance computing software development},
  author={Godoy, William F and Valero-Lara, Pedro and Teranishi, Keita and Balaprakash, Prasanna and Vetter, Jeffrey S},
  journal={Concurrency and Computation: Practice and Experience},
  volume={36},
  number={26},
  pages={e8269},
  year={2024},
  publisher={Wiley Online Library}
}

@misc{valerolara2023comparingllama2gpt3llms,
      title={{Comparing Llama-2 and GPT-3 LLMs for HPC kernels generation}}, 
      author={Pedro Valero-Lara and Alexis Huante and Mustafa Al Lail and William F. Godoy and Keita Teranishi and Prasanna Balaprakash and Jeffrey S. Vetter},
      year={2023},
      eprint={2309.07103},
      archivePrefix={arXiv},
      primaryClass={cs.SE},
      url={https://arxiv.org/abs/2309.07103}, 
}

@article{munley2024llm4vv,
  title={LLM4VV: Developing LLM-driven testsuite for compiler validation},
  author={Munley, Christian and Jarmusch, Aaron and Chandrasekaran, Sunita},
  journal={Future Generation Computer Systems},
  volume={160},
  pages={1--13},
  year={2024},
  publisher={Elsevier}
}

@article{wang2025codevisionary,
  author  = {Xinchen Wang and Pengfei Gao and Chao Peng and Ruida Hu and Cuiyun Gao},
  title   = {{CodeVisionary}: An Agent-based Framework for Evaluating Large Language Models in Code Generation},
  journal = {arXiv preprint arXiv:2504.13472},
  year    = {2025},
}

@inproceedings{zhu2025agenticbenchmarks,
  author    = {Yuxuan Zhu and Tengjun Jin and Yada Pruksachatkun and
               Andy Zhang and Shu Liu and Sasha Cui and Sayash Kapoor and
               Shayne Longpre and Kevin Meng and Rebecca Weiss and
               Fazl Barez and Rahul Gupta and Jwala Dhamala and
               Jacob Merizian and Mario Giulianelli and Harry Coppock and
               Cozmin Ududec and Jasjeet Sekhon and Jacob Steinhardt and
               Antony Kellermann and Sarah Schwettmann and Matei Zaharia and
               Ion Stoica and Percy Liang and Daniel Kang},
  title     = {Establishing Best Practices for Building Rigorous Agentic Benchmarks},
  booktitle = {Advances in Neural Information Processing Systems (NeurIPS),
               Datasets and Benchmarks Track},
  year      = {2025},
}

@article{he2026llmjudge,
  author  = {Junda He and Jieke Shi and Terry Yue Zhuo and Christoph Treude and
             Jiamou Sun and Zhenchang Xing and Xiaoning Du and David Lo},
  title   = {{LLM}-as-a-Judge for Software Engineering: Literature Review, Vision, and the Road Ahead},
  journal = {ACM Transactions on Software Engineering and Methodology},
  year    = {2026},
  note    = {To appear},
}

@inproceedings{tian2024scicode,
  author    = {Minyang Tian and Luyu Gao and Shizhuo Dylan Zhang and
               Xinan Chen and Cunwei Fan and Xuefei Guo and Roland Haas and
               Pan Ji and Kittithat Krongchon and Yao Li and Shengyan Liu and
               Di Luo and Yutao Ma and Hao Tong and Kha Trinh and
               Chenyu Tian and Zihan Wang and Bohao Wu and Yanyu Xiong and
               Shengzhu Yin and Minhui Zhu and Kilian Lieret and Yanxin Lu and
               Genglin Liu and Yufeng Du and Tianhua Tao and Ofir Press and
               Jamie Callan and Eliu Huerta and Hao Peng},
  title     = {{SciCode}: A Research Coding Benchmark Curated by Scientists},
  booktitle = {Advances in Neural Information Processing Systems (NeurIPS),
               Datasets and Benchmarks Track},
  year      = {2024},
}

@inproceedings{xia2026mmscicode,
  author    = {Xue Xia and Zheyuan Yang and Arman Cohan and Yilun Zhao},
  title     = {{MMSciCode}: Real-world Evaluation of Multilingual
               Multi-Discipline Scientific Research Coding},
  booktitle = {Proceedings of the 64th Annual Meeting of the Association for
               Computational Linguistics (Volume 1: Long Papers)},
  pages     = {33981--33999},
  year      = {2026},
}

@article{sun2026scicodepile,
  author  = {Weifeng Sun and Ye Fan and Yuchen Chen and Gou Tan and
             Jieke Shi and Yuan Yidi and Swee Liang Wong and Jonathan Pan and
             David Lo},
  title   = {{SciCodePile}: A 128{GB} Corpus and Executable Benchmark for
             Challenging Scientific Code Generation},
  journal = {arXiv preprint arXiv:2607.19104},
  year    = {2026},
}

@article{lu2026projdevbench,
  title   = {{ProjDevBench}: Benchmarking {AI} Coding Agents on
             End-to-End Project Development},
  author  = {Lu, Pengrui and Zhang, Shiqi and Hou, Yunzhong and
             Ye, Lyumanshan and Huang, Chaoyi and Chen, Zixi and
             Zeng, Ji and Jiang, Hantao and Liu, Pengfei and
             Wang, Yiwei and Yang, Ming-Hsuan},
  journal = {arXiv preprint arXiv:2602.01655},
  year    = {2026}
}

@inproceedings{smith2026improving,
  author = {Smith, Barry and Zhang, Hong and Zhang, Junchao and Balay, Satish and Chen, Le and Ke\c{c}eli, Murat and McInnes, Lois C.},
  title  = {Improving Usability and Productivity of {PETSc} with Agent-Based Workflows},
  booktitle = {PEARC '26: Proceedings of the Practice and Experience in Advanced Research Computing 2026: Resilient Roots + Empowered Communities},
  publisher = {ACM},
  year   = {2026},
  month  = {07},
  note   = {Article 38, \doi{10.1145/3785462.3815882}}
}

\end{document}